\documentclass[10pt,twocolumn,letterpaper]{article}

\usepackage{cvpr}              
\definecolor{cvprblue}{rgb}{0.21,0.49,0.74}
\usepackage[pagebackref,breaklinks,colorlinks,allcolors=cvprblue]{hyperref}
\usepackage{makecell}
\usepackage{multirow}
\usepackage{pifont}
\usepackage{comment}
\usepackage[accsupp]{axessibility}

\usepackage{tabularx}
\newcolumntype{C}[1]{>{\centering\arraybackslash}m{#1}}
\newcolumntype{Y}{>{\centering\arraybackslash}X}
\usepackage{cuted}
\usepackage{multirow}

\title{M3DLayout: A Multi-Source Dataset of 3D Indoor Layouts and Structured Descriptions for 3D Generation}

\author{
Yiheng Zhang$^{1}$\thanks{Equal contribution}\quad
Zhuojiang Cai$^{2}$\footnotemark[1]\quad
Mingdao Wang$^{1}$\footnotemark[1]\quad
Meitong Guo$^{1}$\\
Tianxiao Li$^{1}$\quad
Li Lin$^{3}$\quad
Yuwang Wang$^{1}$\thanks{Corresponding author: wang-yuwang@tsinghua.edu.cn}\\[3pt]
$^{1}$Tsinghua University\quad
$^{2}$Beihang University\quad
$^{3}$Migu Beijing Research Institute\\[3pt]
}

\begin{document}
\maketitle

\begin{strip}
    \centering
    \vspace{-4em}
    \centering
    \includegraphics[width=\textwidth]{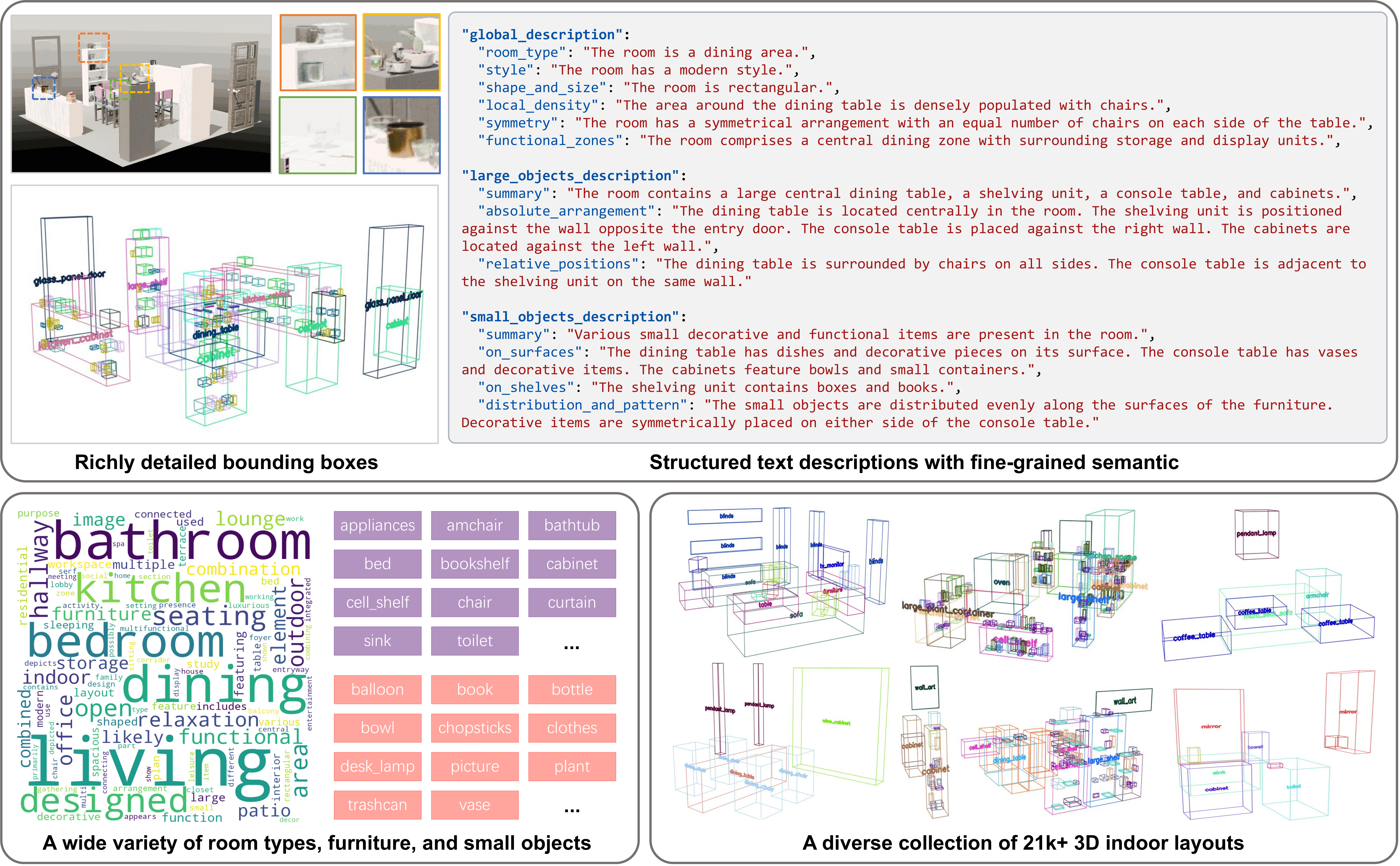}
    \captionof{figure}{\textbf{The M3DLayout dataset — A multi-source benchmark for text-to-3D indoor scene generation.} Top: An example from our dataset showing a detailed 3D indoor layout with richly annotated bounding boxes and its corresponding structured textual description. Bottom-left: Word cloud visualization demonstrating the diversity of room types, furniture, and objects in the dataset. Bottom-right: Overview of the large-scale collection containing 21,367 diverse 3D layout scenes with various styles.}
    \label{fig:teaser}
\end{strip}

\begin{abstract}
In text-driven 3D scene generation, object layout serves as a crucial intermediate representation that bridges high-level language instructions with detailed geometric output. It not only provides a structural blueprint for ensuring physical plausibility but also supports semantic controllability and interactive editing. However, the learning capabilities of current 3D indoor layout generation models are constrained by the limited scale, diversity, and annotation quality of existing datasets. To address this, we introduce \textbf{M3DLayout}, a large-scale, multi-source dataset for 3D indoor layout generation. M3DLayout comprises \textbf{21,367 layouts} and over \textbf{433k object instances}, integrating three distinct sources: real-world scans, professional CAD designs, and procedurally generated scenes. Each layout is paired with detailed structured text describing global scene summaries, relational placements of large furniture, and fine-grained arrangements of smaller items. This diverse and richly annotated resource enables models to learn complex spatial and semantic patterns across a wide variety of indoor environments. To assess the potential of M3DLayout, we establish a benchmark using both a text-conditioned diffusion model and a text-conditioned autoregressive model. Experimental results demonstrate that our dataset provides a solid foundation for training layout generation models. Its multi-source composition enhances diversity, notably through the Inf3DLayout subset which provides rich small-object information, enabling the generation of more complex and detailed scenes. We hope that M3DLayout can serve as a valuable resource for advancing research in text-driven 3D scene synthesis. All dataset and code are released at \url{https://github.com/Graphic-Kiliani/M3DLayout-code}.
\end{abstract}   
\section{Introduction}
\label{sec:intro}

Recent advances in 3D generative modeling have enabled remarkable progress in synthesizing 3D objects and scenes from various modalities, such as text or images. These developments have great potential for downstream applications in areas such as content creation, robotics, and virtual reality~\cite{holleinText2RoomExtractingTextured2023, yangDreamSpaceDreamingYour2024, yangHolodeckLanguageGuided2024a,  schultControlRoom3DRoomGeneration2024}. In particular, text-to-3D generation has attracted increasing attention due to its intuitive and flexible interface for controlling complex scene content.
For example, LucidDreamer~\cite{chungLucidDreamerDomainfreeGeneration2023} and Text2Immersion~\cite{ouyangText2ImmersionGenerativeImmersive2023} adopt point-based or Gaussian-splatting representations to generate detailed scene geometry directly from text or other modalities. Other approaches, such as ATISS~\cite{paschalidouATISSAutoregressiveTransformers2021}, SceneFormer~\cite{wangSceneFormerIndoorScene2021}, EchoScene~\cite{zhaiEchoSceneIndoorScene2025}, and MIDI~\cite{huangMIDIMultiInstanceDiffusion2024}, perform joint layout-object generation by autoregressively predicting room structures and furnishing objects in a unified framework. LayoutGPT~\cite{fengLayoutGPTCompositionalVisual2023}, HoloDeck~\cite{yangHolodeckLanguageGuided2024a}, and InstructScene~\cite{linInstructSceneInstructionDriven3D2024} employ LLMs to plan scene layouts from free-form descriptions, showcasing a promising direction in LLM-driven compositional layout generation.

While these methods demonstrate strong capabilities, they also reveal key limitations. Some of them generate scenes as inseparable volumetric representations, which limits modularity and downstream controllability. Layout-aware models such as CommonScenes~\cite{zhaiCommonScenesGeneratingCommonsense2023} and DiffuScene~\cite{tangDiffuSceneDenoisingDiffusion2024} tend to produce relatively simple scenes with few object types and limited diversity. Meanwhile, LLM-based planners show promise in parsing natural language but often struggle with spatial consistency and accurate physical arrangements. 

To overcome this bottleneck, we draw inspiration from AutoPartGen~\cite{chen2025autopartgen}, which decomposes a scene into distinct parts and thus establishes a bridge between part-level object generation and scene generation. However, AutoPartGen is restricted to small and relatively simple scenes since they simply operate under masked-image conditioning, making it difficult to scale to complex 3D scene environments. In contrast, mainstream object-level part generation methods such as OmniPart~\cite{yang2025omnipart}, X-Part~\cite{yan2025x}, and CoPart~\cite{dong2025copart} rely heavily on 3D bounding boxes as context to control the locations and numbers of parts, highlighting the crucial role of 3D bounding boxes as an intermediate representation. When these 3D bounding boxes are further enriched with semantic labels, rotation information and text descriptions, they form a full 3D layout that not only defines the structural backbone of the scene, but also provides powerful conditional guidance and reflects the functional intent of the environment. Building on this insight, we integrate such layout representations into the scene generation pipeline to produce high-quality, controllable 3D scenes with coherent spatial structure and realistic functionality and further unify object generation and scene generation.

To be more specific, the importance of 3D layouts can be summarized in three key points: $(i)$ 3D layout data define the position, orientation, and scale of objects within a scene, forming the foundation of its structure and spatial coherence. This ensures objects are placed logically and functionally, greatly enhancing the generated scene's realism and credibility while preventing chaotic or illogical visual outcomes. $(ii)$ As powerful conditional information, 3D layout data guides and constrains the 3D scene generation process. It significantly reduces the model's degrees of freedom and ambiguity, allowing for more efficient and accurate detail filling. This also enables users to precisely control the scene's layout, meeting personalized and diverse generation demands.  $(iii)$ Real-world 3D scenes typically serve specific functions, and 3D layout data directly reflect this functionality. Through logical arrangements, generated scenes can better fulfill practical application needs (e.g., interior design, game levels) and provide interactive spaces that align with human habits, thereby significantly improving the end-user experience.

We argue that a major constraint limiting further progress in controllable and high-quality scene generation is the lack of large-scale, richly annotated 3D layout datasets that provide structured, semantic-level supervision. Existing datasets either focus on scene geometry from real-world scans (e.g., ScanNet~\cite{daiScanNetRichlyAnnotated3D2017a}, Matterport3D~\cite{changMatterport3DLearningRGBD2017}) or offer object-level annotations based on professional designs (e.g., 3D-FRONT~\cite{fu3DFRONT3DFurnished2021}, Structured3D~\cite{zhengStructured3DLargePhotorealistic2020}). 
However, none of them offer comprehensive layout annotations—covering large furniture, small objects, and decorative items—paired with structured descriptions of global scene organization and fine-grained spatial relations, and scene feasibility issues, such as overlapping, displacement, and semantic violations, are rarely verified, leaving substantial noise that ultimately limits text-to-layout generation and scene generation.

To address this gap, we introduce M3DLayout, a multi-source dataset for 3D indoor layout generation from structured language. M3DLayout contains 21,367 layouts and over 433k object instances, collected from three complementary sources: real-world scans, professional CAD designs, and procedurally generated scenes.
We pair each layout with structured text descriptions that capture global scene organization, relational placements of large furniture, and fine-grained arrangements of smaller items. These high-quality annotations were constructed using a combination of rule-based extraction, GPT-assisted generation, and rigorous human verification.

To demonstrate the utility of M3DLayout, we establish a benchmark using both a text-conditioned diffusion model and a text-conditioned autoregressive model as baselines for layout generation. Experimental results show that our dataset provides a solid foundation for training layout generation models. Its multi-source composition enhances diversity, notably through the Inf3DLayout subset which provides rich small-object information, enabling the generation of more complex and detailed scenes.

We summarize our main contributions as follows:
\begin{itemize}
    \item We introduce M3DLayout, a large-scale, richly annotated dataset of 3D indoor layouts with structured text descriptions, compiled from diverse complementary sources.
    \item We dedicate efforts to creating the Inf3DLayout subset, which fills a gap in high-quality data for common scene types and substantially increases the level of detail and diversity within the dataset.
    \item We establish two baselines for text-to-layout generation using both diffusion-based and autoregressive baselines. Results show that M3DLayout enhances the diversity and detail of generated scenes, and also enables explicit control over the fineness and density of the generated 3D layout via textual prompts. 
\end{itemize}

\section{Related Work}
\label{sec:Related_Work}

\subsection{Datasets for Indoor Scenes}

Large-scale datasets play a crucial role in learning-based 3D indoor layout generation and scene synthesis. Early efforts to capture real-world environments using 3D scans led to the creation of datasets such as ScanNet~\cite{daiScanNetRichlyAnnotated3D2017a}, Matterport3D~\cite{changMatterport3DLearningRGBD2017}, and SceneNN~\cite{huaSceneNNSceneMeshes2016}. These datasets provide high-fidelity mesh reconstructions, reflect real-world object distributions and spatial constraints. However, they often suffer from noisy geometry, incomplete object coverage, and a lack of fine-grained annotations suitable for generative tasks, as well as limited layout variability due to constrained capture environments.

To address these limitations, synthetic datasets such as SUNCG~\cite{songSemanticSceneCompletion2017}, 3D-FRONT~\cite{fu3DFRONT3DFurnished2021}, and Structured3D~\cite{zhengStructured3DLargePhotorealistic2020} were introduced, offering structured 3D layouts with complete object metadata and annotations from professional CAD designs. However, these professional designs typically lack object variety and fine-grained detail.

Recent hybrid datasets attempt to bridge this gap. FurniScene~\cite{zhangFurniSceneLargescale3D2024} enriches layout realism with more diverse furniture arrangements. OpenRooms~\cite{liOpenRoomsEndtoEndOpen2021} provides photorealistic rendering with physical material properties. However, a key limitation persists across nearly all prior datasets: the lack of scene-level textual annotations, which limits their use for conditional or multimodal generation tasks.


\subsection{Indoor Layout Synthesis}


The evolution of indoor scene synthesis methods highlights the need for richer, more descriptive layout data. Procedural approaches generate indoor scenes using predefined rules, templates, or simulation engines. Systems such as ProcTHOR~\cite{deitkeProcTHORLargeScaleEmbodied2022} and Infinigen~\cite{raistrickInfinitePhotorealisticWorlds2023a, raistrickInfinigenIndoorsPhotorealistic2024} rely on large-scale procedural scene grammars and asset libraries to create diverse, physically realistic environments. These methods offer high controllability and scalability, especially for generating synthetic training data for embodied agents. However, they are ultimately constrained by their handcrafted rules. 

In parallel, a large body of work uses learning-based generative models trained on indoor scene datasets to learn object layouts and spatial relationships in a data-driven fashion. Early methods relied on auto-encoding architectures (e.g., SG-VAE~\cite{purkaitSGVAESceneGrammar2020}, SceneHGN~\cite{gaoSceneHGNHierarchicalGraph2023}, CommonScenes~\cite{zhaiCommonScenesGeneratingCommonsense2023}) and autoregressive models (e.g., ATISS~\cite{paschalidouATISSAutoregressiveTransformers2021}, SceneFormer~\cite{wangSceneFormerIndoorScene2021}). 
More recently, diffusion-based models such as DiffuScene~\cite{tangDiffuSceneDenoisingDiffusion2024} and EchoScene~\cite{zhaiEchoSceneIndoorScene2025} have shown superior performance in capturing complex spatial dependencies. 
SemlayoutDiff~\cite{sun_semlayoutdiff_2025}optimized the Diffuscene~\cite{tangDiffuSceneDenoisingDiffusion2024} by using semantic-mediated 2D distribution maps to finally generate 3D layouts, and is capable of generating different room types with the same model. However, precise control of small object generation cannot be achieved through functional zoning alone.
These methods improve the plausibility and diversity of generated layouts but often face challenges in optimization, attribute disentanglement, and generalization to out-of-distribution scenes.

Inspired by the success of large language models (LLMs), several works explore text-guided 3D scene generation. Architect~\cite{wangArchitectGeneratingVivid2024} and HoloDeck~\cite{yangHolodeckLanguageGuided2024a} leverage world knowledge to interpret user instructions and generate spatial constraints. Similarly, FlairGPT~\cite{littlefair_flairgpt_2025} controls layout generation through more detailed object descriptions using LLMs, but the complex conversational process greatly reduces the efficiency of generating large amounts of data.
While these methods support flexible and intuitive interactions, they often struggle with precise 3D spatial reasoning, often producing physically implausible layouts. This core limitation stems from their lack of grounding in large-scale, physically plausible 3D scene data. M3DLayout is designed to address this critical gap, providing the missing link and a robust foundation to train and benchmark the next generation of these powerful synthesis models.

\begin{figure}[t]
\centering
\includegraphics[width=1.0\linewidth]{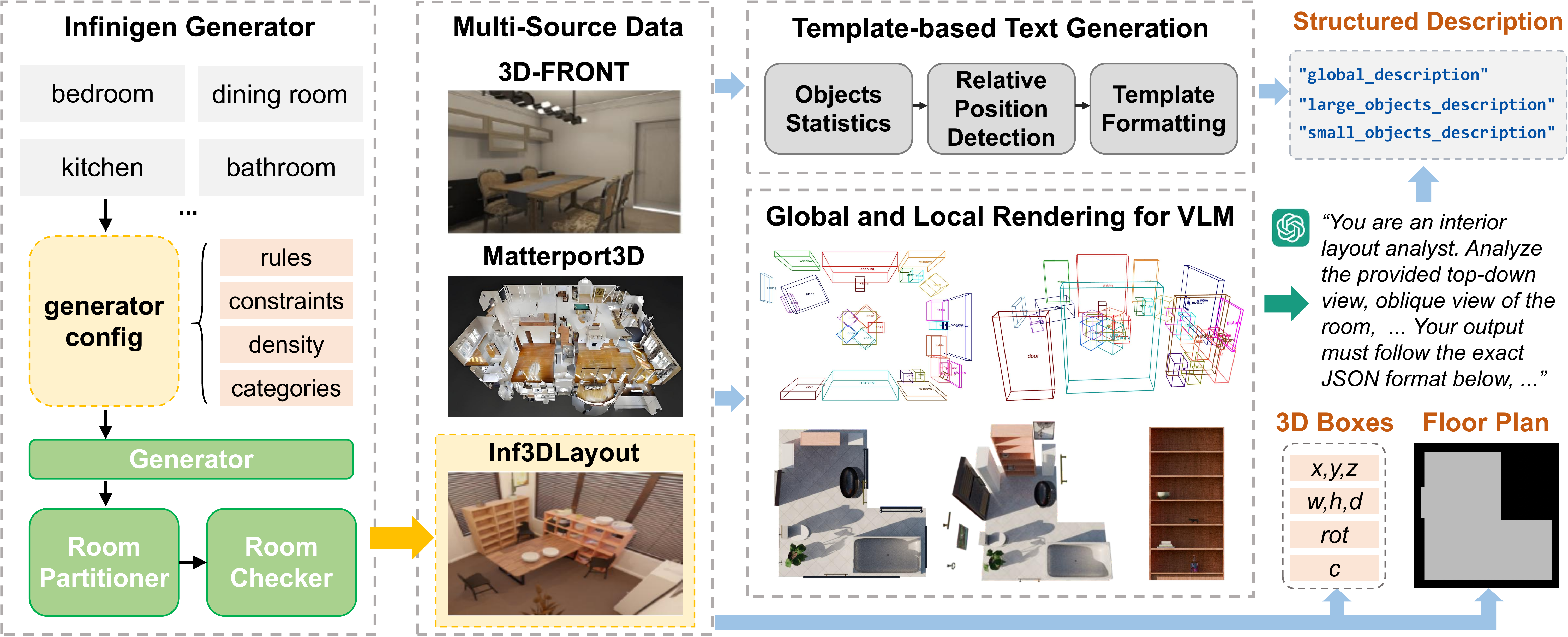}
\caption{\textbf{Pipeline for Constructing the M3DLayout Dataset.} Our framework integrates multi-source data, including the professional designs dataset 3D-FRONT, real-world scans from Matterport3D, and procedurally generated scenes from Infinigen. The construction process involves: meticulously generating, partitioning, and filtering layouts to create the Inf3DLayout subset; performing template-based rules to produce formatted text; and employing global and local rendering for vision-language models (VLM) to produce structured descriptions. This pipeline results in a large-scale, richly-annotated text-3D layout paired dataset.}
\label{fig:pipeline}
\end{figure}

\begin{figure}[t]
\centering
\includegraphics[width=1.0\linewidth]{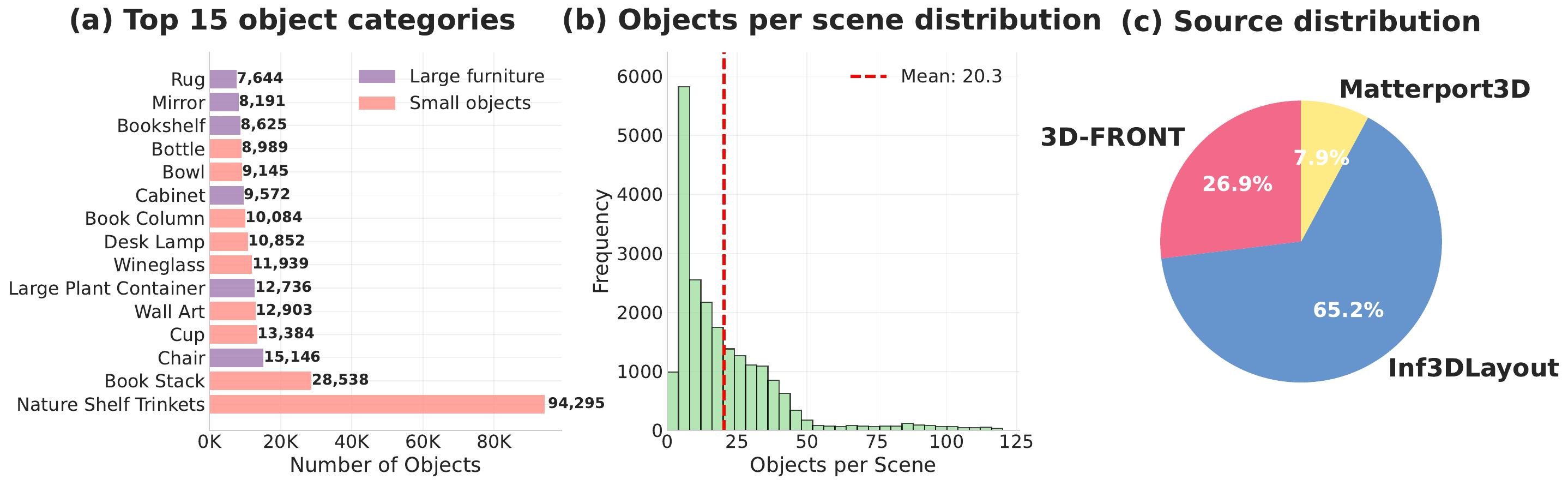}
\caption{\textbf{Dataset statistics of M3DLayout.} (a) Top 15 most frequent object categories. (b) Distribution of the number of objects per scene. (c) Proportion of scenes contributed by each source.}
\label{fig:dataset_statistics}
\end{figure}

\section{The M3DLayout Dataset}
\label{sec:dataset}

To advance controllable and generalizable text-to-3D scene generation, we introduce M3DLayout, a large-scale, multi-source dataset of 3D indoor layouts. This dataset integrates 21,367 layouts from three complementary types of sources: real-world scans, professional interior designs, and procedurally generated scenes. Each layout is annotated with detailed structured textual descriptions to support fine-grained, text-conditioned layout generation.

\subsection{Data Sources and Curation}
\label{sec:data_sources}

M3DLayout is built upon three distinct types of data sources, each contributing unique characteristics:

\noindent\textbf{Real-world Scans.} We incorporate layouts from the Matterport3D dataset \cite{changMatterport3DLearningRGBD2017}, which are derived from real environment scans. These layouts reflect realistic, and often cluttered, spatial arrangements found in actual indoor settings. To ensure data quality, we performed a cleanup of the object category list by merging and removing low-frequency categories. Scenes containing fewer than two object instances were filtered out. This curated subset provides a wide variety of scene types and offers important cues for learning robust and realistic layout patterns.

\noindent\textbf{Professional Interior Designs.} We integrate high-quality layouts from 3D-FRONT \cite{fu3DFRONT3DFurnished2021}, which contains professionally designed indoor scenes. These layouts are characterized by well-organized spatial semantics and adhere to tidy, minimalist design principles. They typically feature sparser object arrangements, providing strong supervision for generating structurally coherent and aesthetically plausible scenes with clean layouts. Following  prior works \cite{paschalidouATISSAutoregressiveTransformers2021, tangDiffuSceneDenoisingDiffusion2024}, we applied filters to remove layouts with uncommon object configurations or unnatural room proportions, ensuring a focus on typical and well-structured  designs.

\noindent\textbf{Procedurally Generated Scenes.} To systematically enhance object diversity, particularly for small and decorative items, we generate the Inf3DLayout subset using the procedural generator Infinigen \cite{raistrickInfinigenIndoorsPhotorealistic2024}. A key part of our curation involved carefully configuring the generator to produce plausible layouts for five common room types: bedrooms, living rooms, dining rooms, kitchens, and bathrooms. The generated houses were then programmatically partitioned into individual rooms. Finally, we applied a filtering step to remove rooms with abnormal layouts or spatial inconsistencies. This curated subset significantly increases the variety and granularity of object arrangements, covering numerous long-tail scenarios that are underrepresented in scan-based or design-based data.

The combination of these sources ensures that M3DLayout encompasses a wide spectrum of indoor environments, balancing realism, design integrity, and compositional diversity.

\subsection{Structured Description Annotation}

To support fine-grained text-conditioned layout generation, we annotate each 3D layout in the M3DLayout dataset with a comprehensive structured description. The annotation schema is designed to capture spatial and semantic information at multiple levels, comprising 3 key components.

\noindent\textbf{Global Scene Description.} This part captures the overall properties and organization of the scene. Each layout is labeled with the room type (e.g., dining area), stylistic attributes (e.g., modern style), and geometric features such as room shape and object density. It also includes high-level functional zoning (e.g., central dining zone with surrounding storage units) and global spatial patterns like symmetry (e.g., equal number of chairs on both sides of the table).

\begin{table}[t]
\centering
\setlength{\tabcolsep}{3pt} 
\footnotesize
\caption{Quantitative analysis of three data sources (3D-FRONT, Matterport3D, Inf3DLayout) in M3DLayout.}
\label{tab:three_sources_comparison}
\resizebox{0.47\textwidth}{!}{%
\begin{tabular}{lrrrrrr}
\toprule
Source       & Scenes & \makecell{Total\\Objects} & \makecell{Avg Objs/\\Scene} & \makecell{Large\\Furniture} & \makecell{Small\\Objects} & Small \% \\
\midrule
3D-FRONT     & 5,754  & 39,494        & 6.9            & 39,407          & 87         & 0.2\% \\
Matterport3D & 1,684  & 21,212        & 12.6           & 12,859           & 8,353         & 39.4\% \\
Inf3DLayout & 13,929  & 373,136       & 26.8           & 117,700          & 255,426       & 68.5\% \\
\midrule
\textbf{} & \textbf{21,367} & \textbf{433,842} & \textbf{20.3} & \textbf{169,966} & \textbf{263,866} & \textbf{60.8\%} \\
\bottomrule
\end{tabular}
}
\end{table}

\noindent\textbf{Large Furniture Description.} We describe the presence and arrangement of major furniture pieces such as dining tables, shelves, consoles, and cabinets. The annotation includes both absolute positioning (e.g., “The shelving unit is placed opposite the entry door”) and relative spatial relations (e.g., “The console table is adjacent to the shelving unit”). A summary of large furniture composition is also provided for each room.

\begin{table}[t]
\centering   
\caption{\textbf{Comparisons between existing 3D indoor scene datasets}. For the column of Layout Collecton, RS denotes real scanned and PD denotes professionally designed .For the column of Variation in
Object Sizes,  “N/A” denotes “not available”, “L” and “S” denote Large and Small objects in the scene, respectively. }
\label{tab:dataset_comparison}
\resizebox{0.479\textwidth}{!}{
\begin{tabular}{lrrccccc}
\toprule
Dataset  & Scenes & Objects &  \makecell{Layout \\ Collection} &  \makecell{Layout \\ Complexity} &  \makecell{Variation in \\ Object Sizes} &  \makecell{Structured \\ Descriptions}  \\
\midrule
 SUN3D \cite{xiao2013sun3d}  & 254     &  N/A   &  RS        & Low        &   N/A   & \ding{55} \\
 SceneNN\cite{huaSceneNNSceneMeshes2016} & 100     &  N/A   &  RS        & Low        &   N/A        & \ding{55} \\
 Matterport3D\cite{changMatterport3DLearningRGBD2017} & 1,684 & N/A &  RS        & Medium         &  L-S         & \ding{55}\\
   ScanNet \cite{daiScanNetRichlyAnnotated3D2017a}    & 1,506 & N/A &  RS        & Low        &  L-S         & \ding{55} \\
    Scan2CAD \cite{avetisyan2019scan2cad}  & 1,506 & N/A & RS         & Low      &   N/A        & \ding{55} \\
    OpenRooms \cite{liOpenRoomsEndtoEndOpen2021} & 1,068 & 97,607& RS        &Low      &   N/A        & \ding{55}\\
\midrule
SceneNet \cite{handa2016understanding}     & 57    &3,699&PD       &Low         & N/A          & \ding{55}\\
Structured3D \cite{zhengStructured3DLargePhotorealistic2020}  &N/A    &N/A  & PD      &Low         & N/A & \ding{55} \\
3D-FRONT \cite{fu3DFRONT3DFurnished2021}      &5,754 &N/A  & PD       &Low      &    L       & \ding{55} \\

\midrule
M3DLayout (Ours)  &21,367 & 433,842 &Mixture          &High        &  L-S     & \ding{51} \\
\bottomrule
\end{tabular}
}
\end{table}

\noindent\textbf{Small Object Description.} This component focuses on decorative and functional small items like dishes, bowls, vases, books, and boxes. These are annotated based on their placement (e.g., on furniture surfaces or shelves) and distribution patterns (e.g., evenly distributed or symmetrically placed). Such fine-grained annotations support more detailed spatial reasoning and realistic scene generation.

The structured descriptions are generated through a multi-stage pipeline, as illustrated in Figure \ref{fig:pipeline}. As outlined in Section \ref{sec:data_sources}, we begin by generating the Inf3DLayout subset using a carefully configured procedural generation pipeline based on Infinigen. For layouts sourced from Matterport3D and the generated Inf3DLayout subset, we render top-down and side views, along with close-up images highlighting the placement of small objects. These multi-view renders are subsequently processed by the GPT-4o model to produce the structured textual descriptions. For scenes from 3D-FRONT, we adopt a rule-based approach: after extracting object-level bounding boxes and semantic labels, we detect relative spatial relationships between objects and format this structured information into coherent descriptions using predefined templates. This methodology is suitable for 3D-FRONT due to the typically simpler and more regular spatial arrangements in its professionally designed scenes, allowing for accurate and comprehensive coverage via template-based generation. Finally, all automatically generated descriptions undergo a sampling-based manual review to ensure annotation quality.

\subsection{Dataset Statistics and Analysis}

M3DLayout encompasses a diverse collection of indoor environments, covering 26 scene categories. Among them, five core room types receive the most focus: bedroom, living room, dining room, kitchen, and bathroom. These constitute the most common residential spaces. In addition to these, the dataset includes functional areas such as office, entryway, closet, toilet, and balcony, as well as specialized spaces including gym, library, and home theater.

\begin{table*}[t]
    \centering
    \caption{\textbf{Quantitative comparison.} Lower FID/KID ($\times$0.001) and higher Clip Score indicate better synthesis quality. FID and KID are computed with respect to the real layouts from 3D-FRONT, Matterport, and Inf3DLayout. We train InstructScene following the public implementation. The optimal result is shown in bold, and the sub-optimal result is shown with an underline.}
    \resizebox{0.9\textwidth}{!}{%
    \begin{tabular}{l|ccccccccc}
        \toprule
        \multirow{2}{*}{\textbf{Method}} & \multicolumn{3}{c}{\textbf{FID ↓}} & \multicolumn{3}{c}{\textbf{KID ↓}} & \multirow{2}{*}{\textbf{CLIP-Score ↑}} \\
        \cmidrule(lr){2-4}\cmidrule(lr){5-7}
         & 3D-FRONT & Matterport & Inf3DLayout & 3D-FRONT & Matterport & Inf3DLayout &\\
        \midrule
        DiffuScene\cite{tangDiffuSceneDenoisingDiffusion2024} & \textbf{29.47} & \underline{98.03}& 102.12&\textbf{10.32} &\underline{47.92} &75.49 &0.1982\\
        InstructScene\cite{linInstructSceneInstructionDriven3D2024} &68.58 &100.54 &159.27 & 54.70&49.23 &156.62 & 0.1944\\
        Ours (DIFF-M3DLayout) & \underline{57.64}& \textbf{87.89}&\underline{70.85} &\underline{36.80} &\textbf{34.62} &\underline{50.94} &\underline{0.2001} \\
        Ours (AR-M3DLayout) & 87.98& 107.58&\textbf{57.90} &57.41 &53.89 &\textbf{41.04} &\textbf{0.2026} \\
        \bottomrule
    \end{tabular}
    }
    \label{tab:room_metrics}
\end{table*}

        

\textbf{Data Composition and Source Characteristics.} As detailed in Table~\ref{tab:three_sources_comparison}, M3DLayout integrates 21,367 layouts with 433,842 object instances, averaging 20.3 objects per scene. The three data sources exhibit complementary characteristics: 3D-FRONT provides professionally designed layouts with clean structural regularity but limited small objects (0.2\%); Matterport3D offers realistic scanned environments with moderate object density (12.6 objects/scene) and a balanced object distribution (39.4\% small objects); while Inf3DLayout significantly enriches the dataset with high scene complexity (26.8 objects/scene) and abundant small objects (68.5\%).

\textbf{Object Distribution and Scene Complexity.} Figure~\ref{fig:dataset_statistics}(a) shows the top 15 most frequent object categories, where small decorative items (e.g., Nature Shelf Trinkets, Book Stack) dominate the distribution, reflecting the dataset's fine-grained annotation richness. The objects-per-scene distribution in Figure~\ref{fig:dataset_statistics}(b) reveals that M3DLayout covers a wide spectrum of scene complexities, from minimalistic arrangements to densely populated environments. This variation enhances the generalization capability of trained models for real-world scenarios.

\textbf{Comparative Advantages.} As shown in Table~\ref{tab:dataset_comparison}, M3DLayout surpasses existing datasets in scale (21,367 scenes, 433k+ objects), layout complexity, and object size variation. Unlike datasets limited to either large furniture (L) or simple layouts, M3DLayout provides comprehensive coverage of both large and small objects (L-S) with structured textual descriptions, addressing a critical gap in current data resources for detailed scene generation.

The broad coverage of room types, the detailed structured descriptions, and the variation in scene complexity make this dataset a valuable resource for downstream tasks such as layout prediction, text-conditioned scene synthesis, and embodied AI simulation.

\section{Indoor Scene Generation}
\subsection{Problem Formulation}

We formulate text-conditioned 3D indoor layout generation as a conditional generative problem. Given a scene-level natural language description $c^{\text{text}}$, the goal is to generate a structured 3D layout $x = \{ o_i \}_{i=1}^N$ consisting of $N$ objects.

Each object $o_i$ is defined as a 3D oriented bounding box:
\begin{equation}
    o_i = (c_i, x_i, y_i, z_i, w_i, h_i, d_i, \theta_i),
\end{equation}
where $c_i$ denotes the semantic class, $(x_i, y_i, z_i)$ is the 3D object center, $(w_i, h_i, d_i)$ are dimensions, and $\theta_i$ is the yaw rotation.  
The objective is to learn a conditional distribution $p_\theta(x \mid c^{\text{text}})$ that captures plausible spatial and semantic relationships among objects given the textual context.

We explore two complementary approaches for modeling this distribution:
(1) a diffusion-based model, which learns to iteratively denoise noisy layout representations toward structured scenes, and  
(2) an autoregressive model, which generates scene objects sequentially in a structured order.  
Both methods are trained to reconstruct layouts consistent with the conditioning text while maintaining spatial coherence and physical plausibility.

\subsection{Diffusion-based Model Architecture}

Our diffusion model follows a denoising diffusion probabilistic framework similar to DiffuScene. The forward process gradually perturbs layout representations $x_0$ with Gaussian noise following a fixed variance schedule $\beta_t$, while the reverse process is parameterized by a neural network $\epsilon_\theta$ that predicts the noise conditioned on the text input:
\begin{equation}
p_\theta(x_{t-1} \mid x_t, c^{\text{text}}) = \mathcal{N}(x_{t-1}; \mu_\theta(x_t, t, c^{\text{text}}), \Sigma_\theta(x_t, t)).
\end{equation}

The model is trained with a noise-prediction objective
\begin{equation}
\mathcal{L}_{\text{DM}} = \mathbb{E}_{x_0, c^{\text{text}}, t, \epsilon} \left[ \| \epsilon - \epsilon_\theta(x_t, t, c^{\text{text}}) \|_2^2 \right],
\end{equation}
and extra regularization terms such as an IoU-based collision penalty to encourage physically valid object arrangements. More details are in the supplementary material.


\subsection{Autoregressive Model Architecture}

To complement the diffusion paradigm, we further employ an autoregressive transformer model that generates scene layouts sequentially.  
Given the textual condition $c^{\text{text}}$, the model factorizes the layout likelihood as
\begin{equation}
p_\theta(x \mid c^{\text{text}}) = \prod_{i=1}^{N} p_\theta(o_i \mid o_{<i}, c^{\text{text}}),
\end{equation}
where each object $o_i$ is predicted conditioned on the previously generated objects and the text.

The autoregressive model is implemented with a transformer encoder architecture. Text tokens and previously generated object embeddings form a unified input sequence for context-aware sequential prediction. The training objective minimizes the negative log-likelihood of object parameters under the predicted conditional distribution. Further details are provided in the supplementary material.
\section{Experiments}
\subsection{Experimental Settings}
\begin{figure*}[t]
    \centering
    \includegraphics[width= 0.8\linewidth]{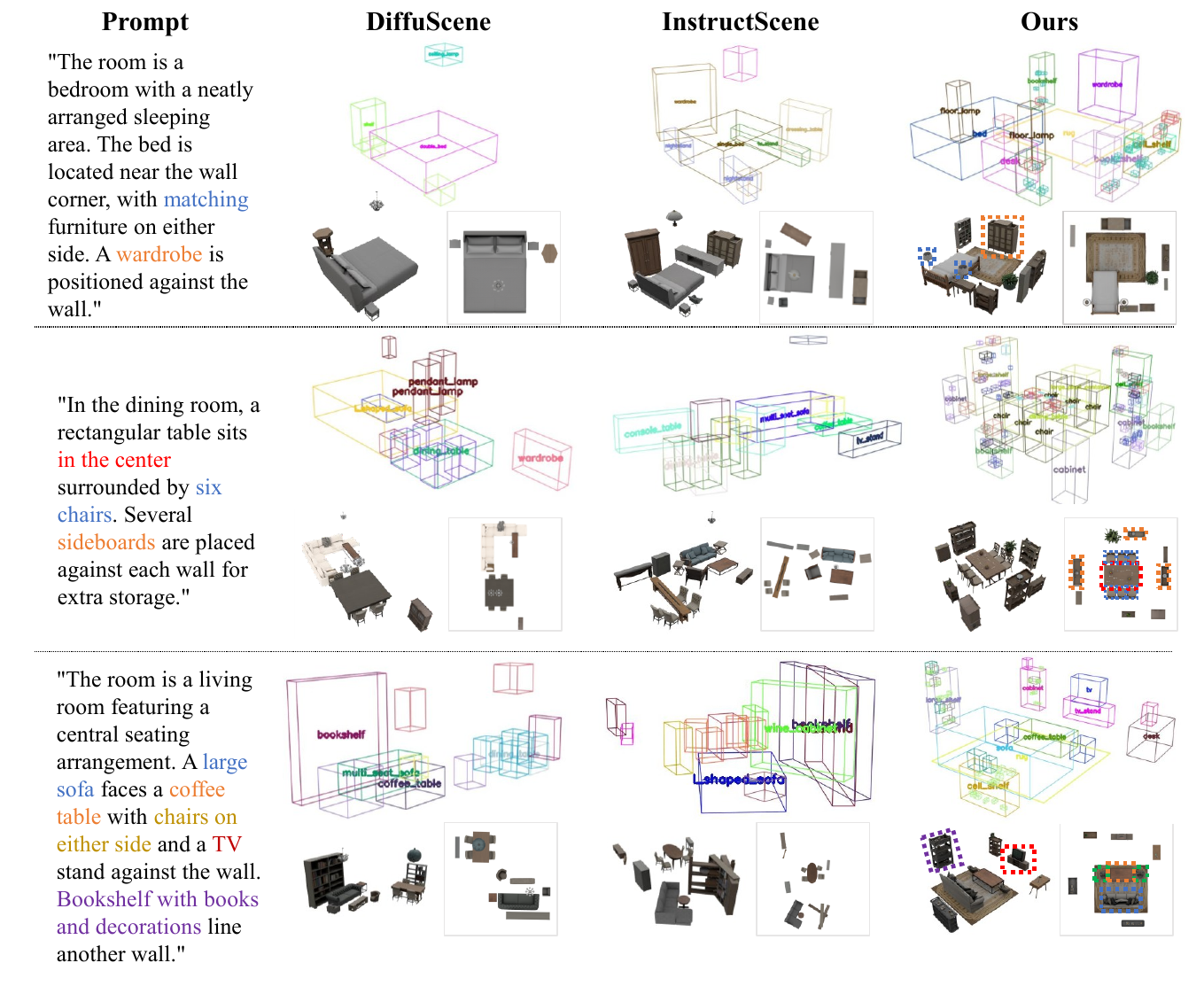}
    \caption{
        \textbf{Qualitative comparison of different methods on diverse room types.}
        From top to bottom: bedroom, dining room, and living room generation results.
        Each row shows the input prompt and generated layouts from DiffuScene, InstructScene, and our method.
        Trained on the M3DLayout dataset, our method produces richer layout details from text descriptions.
    }
    \label{fig:qualitative_results}
\end{figure*}

We evaluate the effectiveness of autoregressive and diffusion-based methods on the proposed dataset, assessing the plausibility and controllability of generated 3D layouts. Detailed train/val splits and implementation details are in the supplementary material.



\textbf{Baselines.} 
We compare our method with two state-of-the-art scene generation approaches: DiffuScene~\cite{tangDiffuSceneDenoisingDiffusion2024} and InstructScene~\cite{linInstructSceneInstructionDriven3D2024}, both of which are text-driven methods based on diffusion models. More details are provided in the supplementary material.



\textbf{Metrics.} 
Following prior works~\cite{tangDiffuSceneDenoisingDiffusion2024, linInstructSceneInstructionDriven3D2024}, we adopt \text{Fr\'echet} Inception Distance (FID)~\cite{heusel2017gans} and Kernel Inception Distance (KID)~\cite{binkowski2018demystifying}, and the CLIP-Score~\cite{hessel2021clipscore} to measure the fidelity and controllability, respectively. We employ a self-designed object retrieval method to realize instance filling from layout to scene and rendering. We also calculate more metrics to measure the complexity and diversity of our dataset. More details are provided in the supplementary material. 


\begin{figure*}[t]
    \centering
    \includegraphics[width=\linewidth]{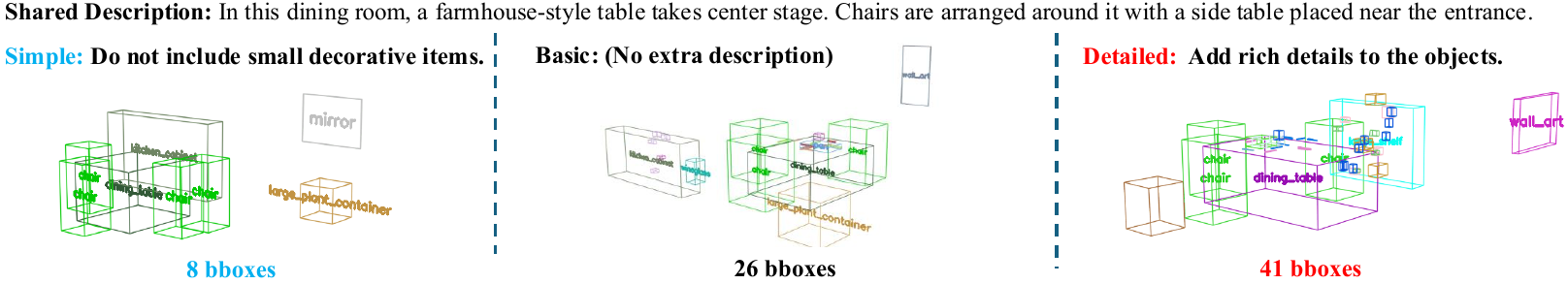}
    \caption{
        \textbf{Density controllability in layout generation with different input texts.}
        The first row presents input prompts for our layout generation model, showcasing variations in objects density from low to high, with minor changes in the last sentence. The second row illustrates the corresponding output results generated by our model, which adapt based on the prompt density. 
    }
    \label{fig:small_object_density_controllability}
\end{figure*}

\subsection{Main Results}
\textbf{Quantitative Comparison.}

The quantitative comparison results are shown in Table~\ref{tab:room_metrics}, where ``DIFF-M3DLayout" and ``AR-M3DLayout" represent diffusion- and autoregressive-based methods, respectively. In Table~\ref{tab:room_metrics}, the horizontal axis represents ground-truth renderings (real images) from three different datasets. The same set of 1,500 prompts is used as conditioning input for all comparative methods to generate layouts, which are then rendered as synthetic images and used together with real images to compute FID and KID. More details can be found in the supplementary material.

As demonstrated, for FID and KID, the proposed DIFF-M3DLayout drastically outperforms the state-of-the-art, achieving improvements of 10\%–32\% on the reference Matterport dataset and Inf3DLayout datasets, and AR-M3DLayout achieves 34\%–44\% improvements on Inf3DLayout dataset, which demonstrates superior generalization compared with DiffuScene and InstructScene.
On 3D-FRONT, however, it falls behind DiffuScene on these metrics. This is primarily because the number of objects per scene in 3D-FRONT typically ranges from 5 to 12, whereas our method generates scenes with more than 12 objects in most cases, causing a mismatch in scene complexity distribution. As a result, the visual statistics of our generated layouts deviate more from the ground truth (3D-FRONT), which negatively impacts FID and KID. Conversely, the richer object counts and variety in the scenes generated by our model provide another advantage of our method relative to the multi-source datasets M3DLayout, which is also confirmed by the visualizations in Figure~\ref{fig:qualitative_results}. Moreover, our method surpasses the baselines in CLIP-Score, demonstrating enhanced controllability and a stronger alignment between the generated scenes and the given prompts.


\textbf{Qualitative Comparison.}
The proposed DIFF-M3DLayout is employed for qualitative comparison with diverse methods, with the results in the bedroom, dining room, and living room shown in Figure~\ref{fig:qualitative_results}.
Across all settings, the proposed DIFF-M3DLayout demonstrates improved semantic controllability and visual fidelity compared to Diffuscene and Instructscene. Specifically, DiffuScene can generate scenes that appear visually neat at first glance but struggles to produce small objects and exhibits limited prompt controllability, as illustrated by the living room case. InstructScene, on the other hand, fails to accurately model spatial relationships among instances, leading to disorderly object placement, as shown in the dining room case. By contrast, the proposed DIFF-M3DLayout generates precise and diverse objects in accordance with the prompt semantics—such as the small items on the shelf and six chairs in the dining room case—while effectively capturing 3D spatial arrangements, exemplified by ``The bed is located near the wall corner" in the bedroom case.

We perform the ablation experiments using the diffusion-based method to validate the effectiveness of a single training dataset. See more details in the supplementary material.

\subsection{Density Controllability}
We verify the controllability of object density in our layout generation model using different input prompts. As shown in Figure~\ref{fig:small_object_density_controllability}, the first row displays the different types of input prompts, which only differ in the final sentence. These prompts adjust the density of objects within the layout, from a minimal setup to one with richer details. 
This highlights the model’s ability to control scene density based on the granularity of the input description, thus offering flexibility in layout customization for various applications.

\subsection{User Case Study}

\begin{figure}[t]
\centering
\includegraphics[width=1.0\linewidth]{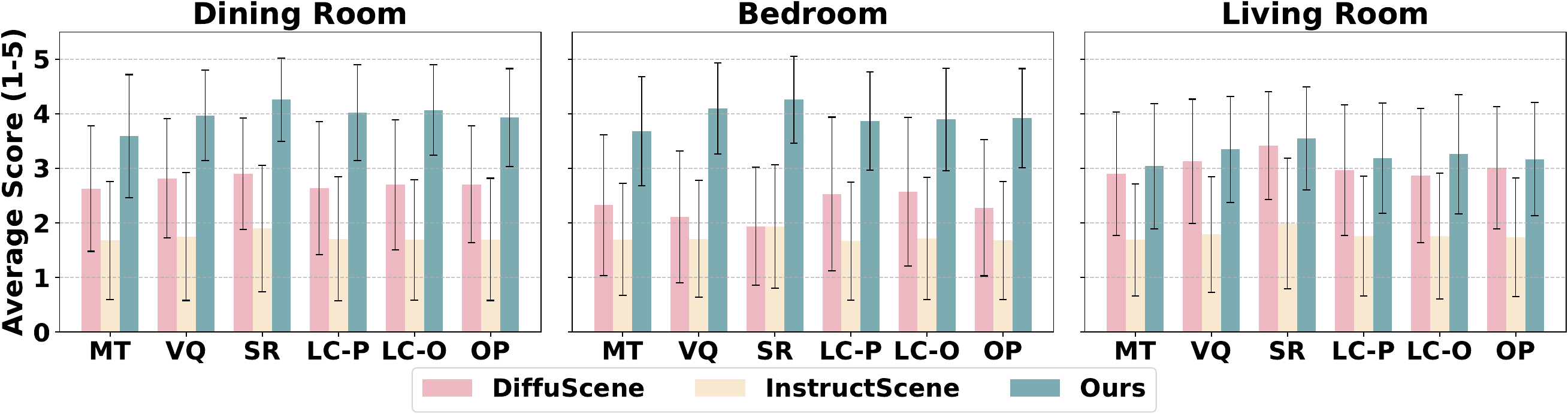}
\caption{\textbf{User case study results.} The charts compare our method against DiffuScene and InstructScene across diverse rooms. Bars represent the average user score for six metrics: Match with Text (MT), Visual Quality (VQ), Scene Richness (SR), Layout Coherence (Position) (LC-P), Layout Coherence (Orientation) (LC-O), and Overall Preference (OP).}
\label{fig:user_case_study}
\end{figure}

We conducted a perceptual study to evaluate the quality of our generated layouts against DiffuScene and InstructScene. We recruited 42 participants to rate 15 scenes across three room types (dining room, bedroom, and living room). For each scene, participants were shown a text description, a top-down rendering, and the generated layout for each of the three methods, rating them on a 1-to-5 scale across six metrics. All participants were volunteers. The overall results are summarized in Figure~\ref{fig:user_case_study}. Our method outperformed both baselines in the majority of metrics and room categories. The most significant advantage was observed in the Scene Richness metric. The performance gap was less pronounced in living rooms, likely because key layouts are harder to assess from the top-down view compared to the iconic beds or dining sets in other rooms. These results confirm that users perceive our generated layouts as more detailed, coherent, and of higher quality.

\section{Conclusion}

We introduced M3DLayout, a large-scale, multi-source dataset for 3D indoor layout generation from structured text descriptions. It integrates real-world scans, professional designs, and procedurally generated scenes. Each layout is paired with structured descriptions that cover global scene summaries, relational placements of large furniture, and fine-grained arrangements of small objects. This multi-source, richly annotated structure enables the learning of diverse spatial and semantic patterns across a wide variety of indoor environments. We established a benchmark using both a text-conditioned diffusion model and an autoregressive model, and experimental results validate that training on our dataset enhances the diversity and detail of generated layouts, with the Inf3DLayout subset in particular enabling more complex and richly annotated scenes. We hope M3DLayout serves as a valuable resource for advancing research in text-driven 3D scene synthesis.

\section{Acknowledgement}
This work was supported by National Science and Technology Major Project (No. 2022ZD0119404), and the National Natural Science Foundation of China (No. 62576184).

\clearpage
\setcounter{page}{1}
\maketitlesupplementary


\section{Supplementary Material Overview} 
This supplementary document provides additional technical details and visualization results to support the main paper. The supplementary is organized into four sections:

\paragraph{\cref{sec:implementation_details} Implementation Details.} We provide detailed explanations of the dataset splitting strategy in ~\ref{sec:datasplit}, both diffusion and autoregressive based model details in ~\ref{sec:modeldetails}, training details in ~\ref{sec:trainingdetails}, compared baselines details in ~\ref{sec:baseline} and metric details in ~\ref{sec:metrics}. 

\paragraph{\cref{sec:analysis} More Dataset Analysis.} We further evaluate the quality of our dataset through extensive analysis and validation. Specifically, we quantitatively compare more metrics with 3D-FRONT in ~\ref{sec:complexity}, conduct a human evaluation via a dedicated website in ~\ref{sec:humanevaluation} to demonstrate its overall complexity, diversity and high quality,.

\paragraph{\cref{sec:visualization} More Visualization Results.} We provide additional examples that further demonstrate the quality of our dataset, the fidelity of our generative model, and its coherence with textual inputs.

\paragraph{\cref{sec:ablation} Ablation.} We conduct some ablation studies to show that models trained on a single dataset overfit to its distribution and generalize poorly, whereas training on the multi-source M3DLayout dataset yields consistently balanced, realistic, and controllable layouts across diverse data types.

\paragraph{\cref{sec:objectretrieval} Object Retrieval.}We present a scalable and self-developed layout-to-scene object retrieval pipeline that accurately maps generated layouts to 3D assets, facilitating reliable metric evaluation and high-fidelity visualizations.

\section{Implementation Details}

\label{sec:implementation_details}


\subsection{\textbf{Dataset Split}}
\label{sec:datasplit}
Our experiments are conducted on a subset of the M3DLayout dataset containing 15,080 scenes. This subset is randomly divided into 12,062 layouts for training and 3,018 layouts for validation. For the ablation studies in Table~\ref{tab:ablation}, models are additionally trained on three independent datasets with the following splits (training/validation): 4,603/1,151 for 3D-FRONT, 1,347/337 for Matterport3D, and 6,112/1,530 for Inf3DLayout.
For all evaluations, and to ensure a fair comparison with prior work, we do \emph{not} use the scene description texts provided in M3DLayout. Instead, we generate 500 scene descriptions each for \emph{bedroom}, \emph{dining room}, and \emph{living room} using GPT-4o, resulting in a total of 1,500 test descriptions. This unified test set is used across all experiments for fairness.

\subsection{Model Details}
\label{sec:modeldetails}
\noindent\textbf{Diffusion-based Model.}
Our diffusion-based model represents a scene as a sequence of $N$ objects, where the maximum number of objects $N$ is fixed at 120. Padding objects with a $\mathrm{PAD}$ category label is applied to maintain a consistent sequence length. The sequence is ordered by the $(x, z, y)$ coordinates of the bounding box centers, following recent best practices in the 3D part-level generation community for layout generation. Each object is parameterized by a concatenated vector 
\(
[l_i, s_i, \cos\theta_i, \sin\theta_i, c_i],
\)
which includes its location ($l_i \in \mathbb{R}^3$), size ($s_i \in \mathbb{R}^3$), orientation ($\theta_i \in \mathbb{R}$, encoded using $\cos\theta_i$ and $\sin\theta_i$), and category label ($c_i \in \mathbb{R}^C$). 

The denoiser is a UNet-based architecture built upon 1D convolutions with skip connections. It independently encodes and predicts the distinct attributes of the object. We use a pre-trained BERT encoder to extract text embeddings $z$ from the input description. This language guidance is injected into the denoiser network through cross-attention layers, enabling the network to predict the noise $\epsilon_{\phi}$ conditioned on the timestep $t$ and the text embedding $z$.


The training objective consists of the scene loss $L_{\mathrm{sce}}$, which minimizes the difference between the actual and predicted noise, supplemented with an IoU regularization term $L_{\mathrm{iou}}$. For two 3D axis-aligned bounding boxes $A$ and $B$, we define
\[
L_\mathrm{IoU}(A,B) = \frac{I}{V_A + V_B - I + \epsilon},
\]
\[
I = \prod_{k \in \{x,y,z\}} \max\!\left(0,\min\!\left(k_2^A,k_2^B\right)-\max\!\left(k_1^A,k_1^B\right)\right).
\]
This term penalizes object intersections and encourages physically plausible arrangements.

The key difference between DiffuScene and Our DIFF-M3DLayout is the use of shape codes (object geometric embedding) during training. We trained DIFF-M3DLayout (our diffusion-based trained on M3DLayout) only with layout data (without shape codes) because a subset of M3DLayout, Matterport, comes from real-world scans and lacks complete object shapes to extract shape codes. Table~\ref{diffcomparison} shows metrics of the two methods trained on 3D-FRONT. The performance is quite similar.


\begin{table*}[t]
    \centering
    \caption{{DiffuScene and Our DIFF-M3DLayout.} }
    \vspace{-3mm}
    \resizebox{1.0\textwidth}{!}{%
    \begin{tabular}{l|ccccccccc}
        \toprule
        \multirow{2}{*}{\textbf{Method}} & \multicolumn{3}{c}{\textbf{FID ↓}} & \multicolumn{3}{c}{\textbf{KID ↓}} & \multirow{2}{*}{\textbf{CLIP-Score ↑}} \\
        \cmidrule(lr){2-4}\cmidrule(lr){5-7}
         & 3D-FRONT & Matterport & Inf3DLayout & 3D-FRONT & Matterport & Inf3DLayout &\\
        \midrule
        DiffuScene (trained on 3D-FRONT) & 29.47 & 98.03& \textbf{102.12}&\textbf{10.32} &47.92 &\textbf{75.49} &0.1982\\
        Ours diffusion-based trained on 3D-FRONT & \textbf{27.33}&\textbf{83.88} &110.98 &10.59 &\textbf{21.80} &83.45 &\textbf{0.2083} \\
        
        \bottomrule
    \end{tabular}
    }
    \label{diffcomparison}
\end{table*}

\noindent\textbf{Autoregressive Model.}
Our autoregressive model shares the same scene representation as our diffusion-based model for consistency. The autoregressive approach does not have a strict upper limit on the sequence length, but we still utilize data containing up to 120 objects per scene for training. A scene is represented as a sequence of objects, bookended by special start-of-sequence ($\mathrm{SOS}$) and end-of-sequence ($\mathrm{EOS}$) tokens.

The input text description is similarly encoded into text embeddings using a pre-trained BERT model. At each generation step, the input sequence to the model consists of the text token embeddings and the embeddings of all previously generated objects. A Transformer encoder serves as the backbone network. The attributes of each previously generated object are first projected by an MLP into an embedding space. The network then attends to this combined sequence to predict the parameters of the next object.

The training objective is the scene loss $L_{\mathrm{sce}}$, which minimizes the negative log-likelihood of the ground-truth object parameters given the input context.

\subsection{Training Details.}
\label{sec:trainingdetails}
Both our diffusion and autoregressive based models are trained for 30k epochs using the Adam optimizer. For the diffusion model, we use a learning rate of $2 \times 10^{-4}$ with a step-wise decay factor of 0.5 every 10k steps and a linear noise schedule. For the autoregressive model, we use a learning rate of $1 \times 10^{-4}$.




\begin{figure}[h]
\centering
\includegraphics[width=1.0\linewidth]{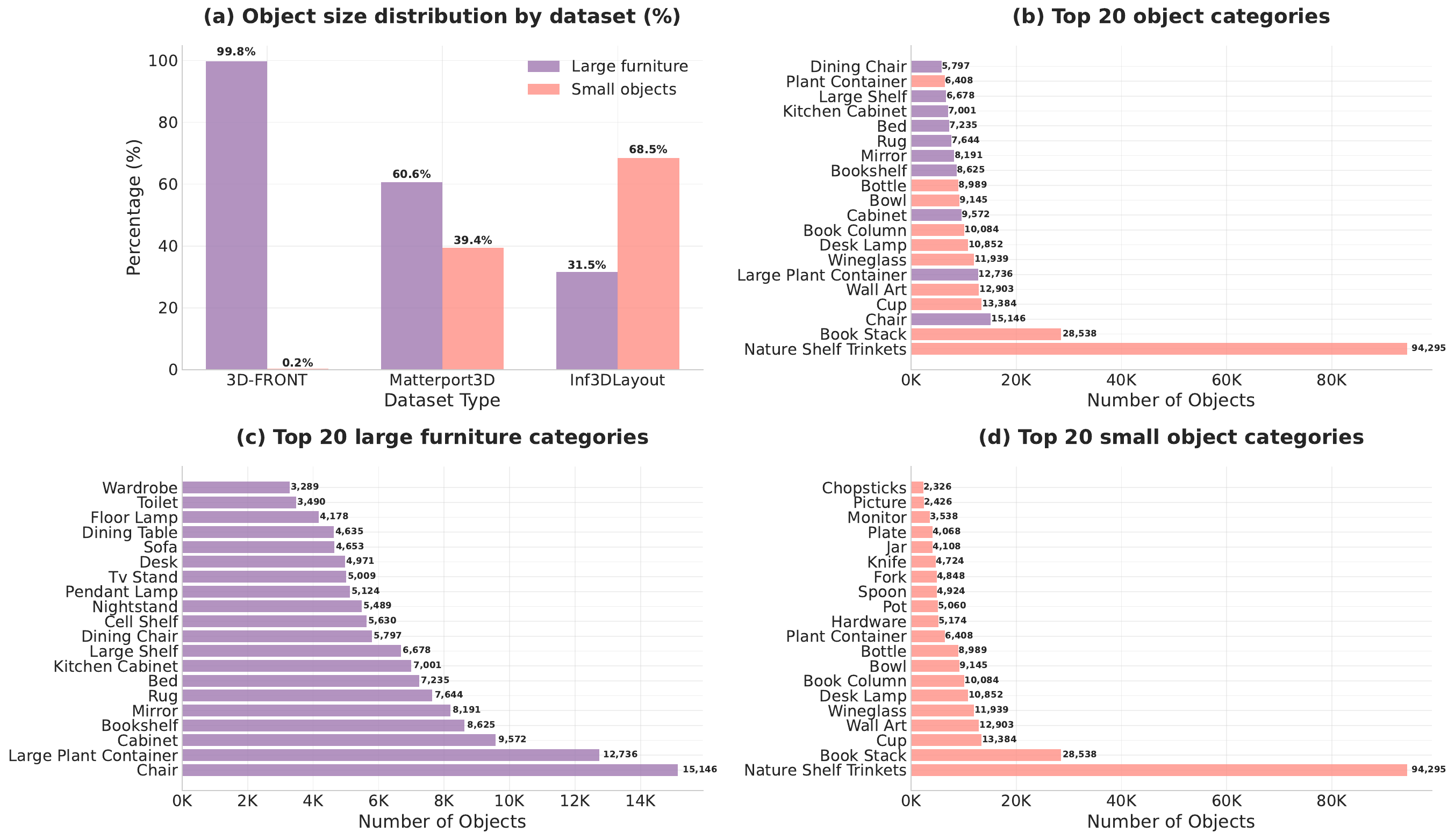}
\caption{\textbf{Object distribution statistics of the M3DLayout dataset.} (a) Size distribution (large/small) of objects by source. (b) Overall ranking of the top 20 object categories. (c-d) Rankings for large and small objects, respectively.}
\label{fig:cat_analyisis}
\end{figure}

\subsection{Compared Baselines}
\label{sec:baseline}
We compare our method with two state-of-the-art scene generation approaches: (1) DiffuScene~\cite{tangDiffuSceneDenoisingDiffusion2024}, a diffusion model for 3D indoor scene synthesis that denoises unordered object attributes to produce physically plausible layouts. (2) InstructScene~\cite{linInstructSceneInstructionDriven3D2024}, a graph diffusion model that integrates a semantic graph with a layout decoder to synthesize 3D indoor scenes from natural language instructions. Both methods allow for the conditioning on text prompts. For inference with DiffuScene, we employ the officially released model weights for the bedroom, dining room, and living room, and follow the public implementation to train InstructScene on the same room types.

\subsection{Metrics}
\label{sec:metrics}
Following prior works~\cite{tangDiffuSceneDenoisingDiffusion2024, linInstructSceneInstructionDriven3D2024}, we adopt \text{Fr\'echet} Inception Distance (FID)~\cite{heusel2017gans} and Kernel Inception Distance (KID)~\cite{binkowski2018demystifying} to quantify the fidelity of scenes synthesized from layouts by measuring the similarity between generated and ground-truth top-down renderings. Meanwhile, we employ the CLIP-Score~\cite{hessel2021clipscore} to evaluate the controllability of generated layouts by computing the cosine similarity between CLIP-encoded features of the generated renderings and the given prompts. To this end, we first employ a text2mesh model~\cite{xiang2025structured} to generate the required object instances and then retrieve both the ground-truth and synthesized scenes conditioned on the layout.

As for the calculation of FID and KID, the real images are typically taken from the training set; however, since our method and the baselines are trained on different datasets with varying data distributions, direct comparison would be unfair. To address this, we select different datasets as the source of real images, allowing for both a fair comparison and an evaluation of fidelity and controllability.

\begin{table*}[t]
\caption{Comparison of generation efficiency with state-of-the-art methods.}
\label{tab:efficiency}
\centering
\begin{tabular}{lccc}
\toprule
Method & \makecell{Generation Time \\ per Scene (s)} & \makecell{Generation Time \\ per BBox (s)} & \makecell{Average BBoxes \\ per Scene} \\
\midrule
Diffuscene & 17.4067 & 1.7674 & 9.849 \\
InstructScene & \underline{1.3438} & \underline{0.1280} & 10.502 \\
Ours (DIFF-M3DLayout) & 30.18 & 1.639 & \textbf{18.4} \\
Ours (AR-M3DLayout) & \textbf{0.2348} & \textbf{0.01587} & \underline{14.8} \\
\bottomrule
\end{tabular}
\end{table*}

\section{More Dataset Analysis}
\label{sec:analysis}

\subsection{Layout Complexity And Diversity}
\label{sec:complexity}
We further strengthened our quality validation by quantitatively benchmarking our dataset. We evaluated the data across key dimensions, including scale, layout complexity, and diversity. The quantitative results in Table~\ref{tab:quality_comparison} demonstrate that our M3DLayout significantly surpasses 3D-FRONT across all these metrics, confirming the higher quality and richness of our data. To provide a robust assessment, we evaluate layout complexity using the Average Number of Large Objects (identified by categories, to ensure independence from small-object density) and Average Token Counts (BPE sequence length) as a proxy for description to model the layout. For dataset diversity, we benchmarked 500 scenes using Category Entropy (measuring object class uniformity), Intra-Category Spatial Distance (calculating average weighted nearest-neighbor distance for geometric variation), and CLIP Embedding Distance (averaging cosine distances of text embeddings for semantic diversity, $D_{\text{text}}=\frac{1}{|\mathcal{P}|}\sum_{(i,j)\in\mathcal{P}}\bigl(1-\langle \hat{\mathbf e}_{i},\hat{\mathbf e}_{j}\rangle\bigr)$).

\subsection{Human Evaluation}
\label{sec:humanevaluation}
Besides User Case Study, we conduct a human evaluation on 1,000 (4.68\% of total) randomly sampled scenes from M3DLayout. 
Evaluators (around 200) are asked to rate three aspects: layout rationality, overall description accuracy, and object-level description accuracy, using a 5-point Likert scale (5: completely correct, 1: completely incorrect). 
{Each sample is independently evaluated by two participants, with cross-validation applied to ensure consistency.}
The results show that the average scores for the three criteria are 4.423, 4.462, and 4.436, respectively. 
Moreover, the proportions of samples receiving scores $\geq 4$ are 94.05\%, 94.08\%, and 93.84\%, respectively. 
These results further validate the high quality of the M3DLayout dataset and demonstrate its strong alignment with human judgments

\begin{table*}[h]
    \centering
    \caption{
    M3DLayout with 3D-FRONT comparison. 
    }
    \vspace{-3mm}
    \label{tab:quality_comparison}
    \resizebox{\linewidth}{!}{
        \begin{tabular}{l | c c c | c c c}
        \toprule
        & \multicolumn{3}{c|}{\textbf{Layout Complexity}} & \multicolumn{3}{c}{\textbf{Dataset Diversity}} \\
        Dataset & Avg. \# Objs & Avg. \# Large & Avg. Tokens & Entropy $\uparrow$ & Spatial Dist. $\uparrow$ & CLIP Dist. $\uparrow$ \\
        \midrule
        3D-FRONT & 6.9 & 6.8 & 19.5 & 0.635 & 1.416 & 0.133 \\
        M3DLayout (Ours) & \textbf{26.8} & \textbf{8.4} & \textbf{77.9} & \textbf{0.720} & \textbf{3.854} & \textbf{0.203} \\
        \bottomrule
        \end{tabular}
    }
\end{table*}

\section{More Visualization Results}
\label{sec:visualization}
\subsection{Dataset Layout Visualization}
\begin{figure*}[h]
\centering
\includegraphics[width=1.0\textwidth]{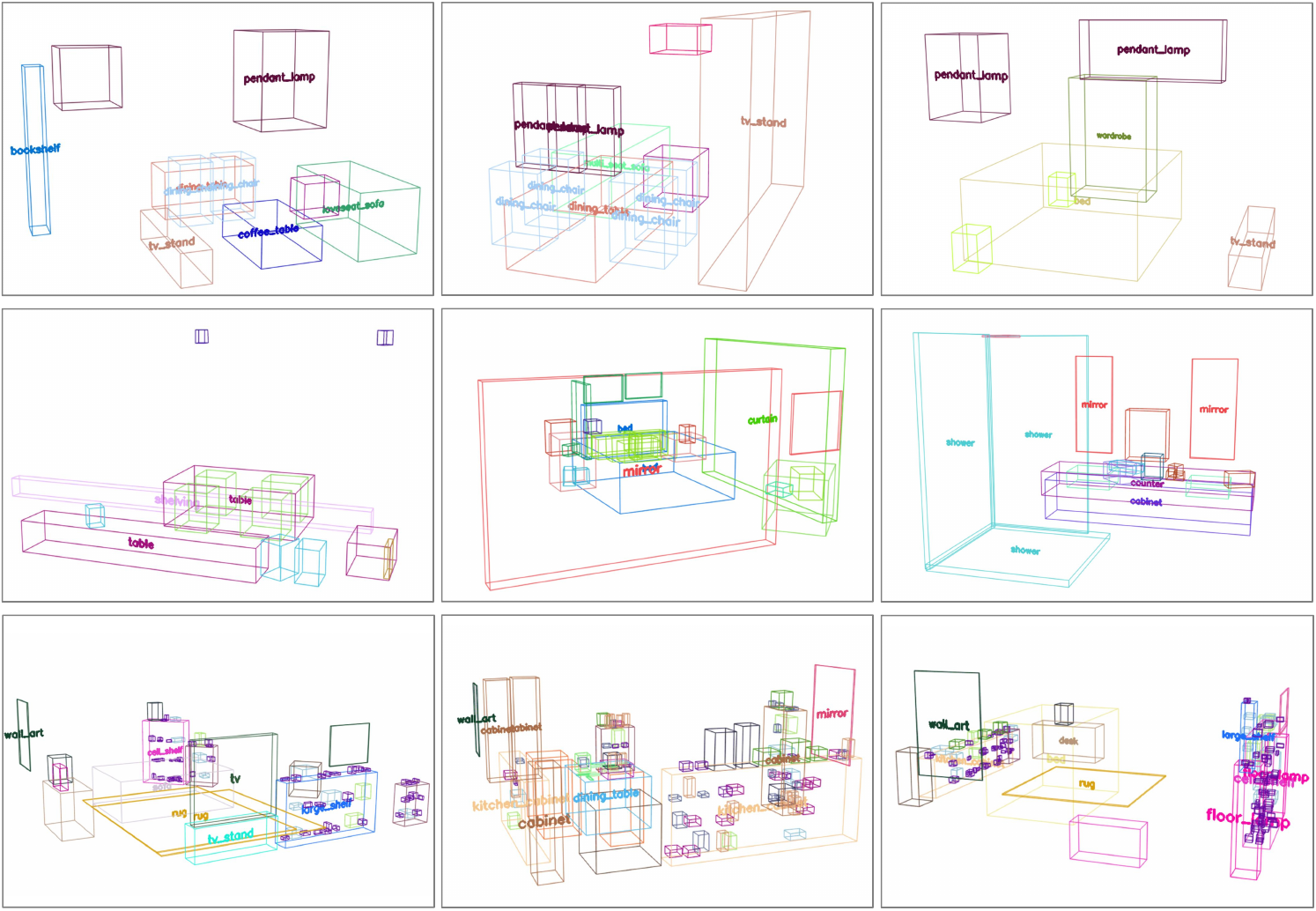}
\caption{Samples from the M3DLayout dataset. The first row shows scenes from CAD designs (3D-FRONT), the second row from real-world scans (Matterport3D), and the third row from procedurally generated scenes (Infinigen).}
\label{fig:dataset_sample}
\end{figure*}

We provide diverse types of 3D scene data from our M3DLayout dataset in Figure~\ref{fig:dataset_sample}, which includes scenes from CAD designs sourced from the 3D-FRONT dataset at the first row, scenes derived from real-world scans, specifically from the Matterport3D dataset at the second row and procedurally generated scenes  Inf3DLayout from Infinigen at the third row. These images demonstrate the flexibility of the M3DLayout dataset in representing a wide spectrum of interior environments, from synthetic CAD designs to real-world captures and generative models.

\subsection{Generated Layout Visualization}
\begin{figure*}[h]
\centering
\includegraphics[width=1.0\textwidth]{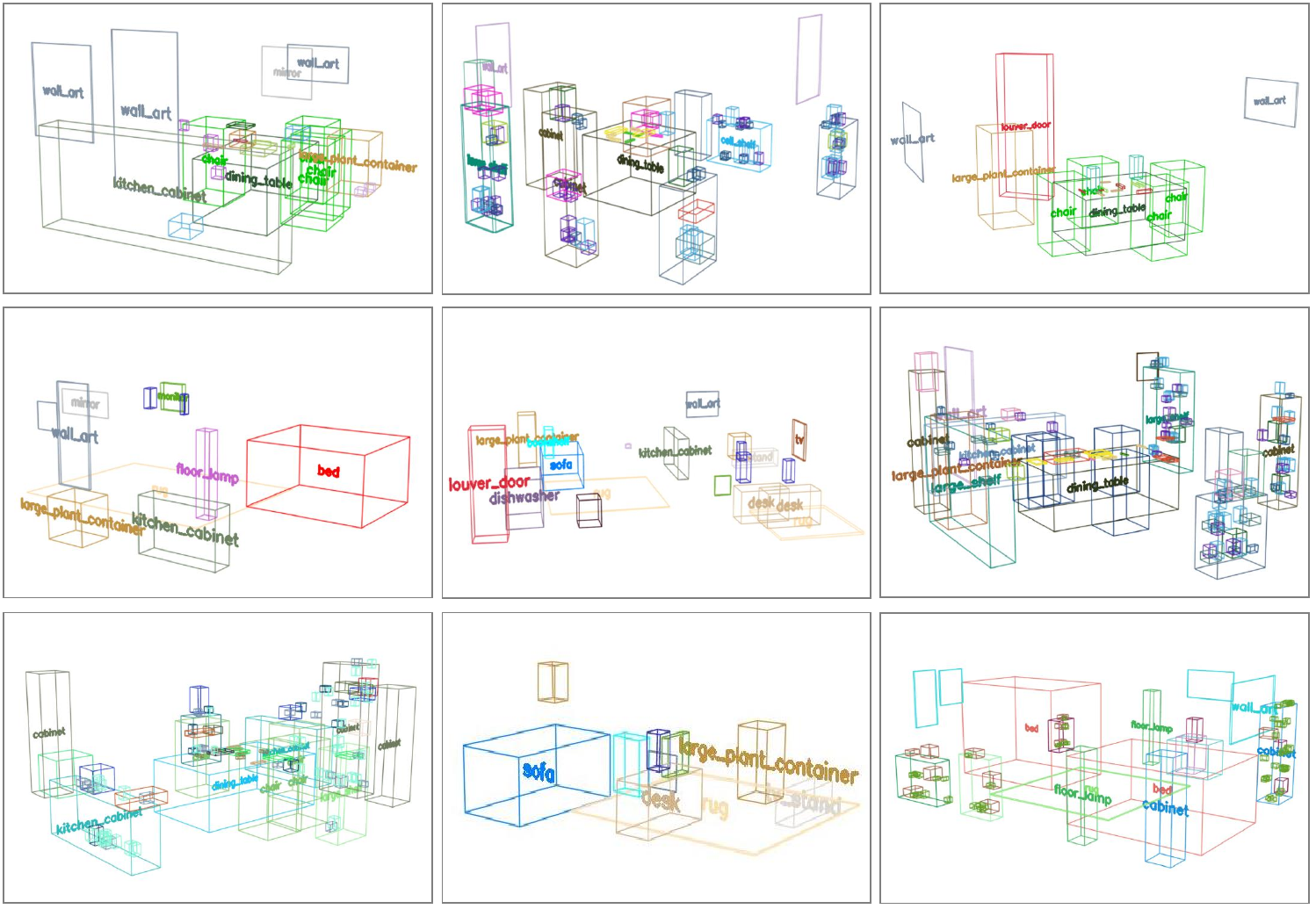}
\caption{Generated layouts by our model trained on the M3DLayout dataset, visualized for randomly selected bedroom, living room, and dining room.}
\label{fig:fig_dataset_generation}
\end{figure*}

\begin{figure*}[h]
\centering
\includegraphics[width=1.0\textwidth]{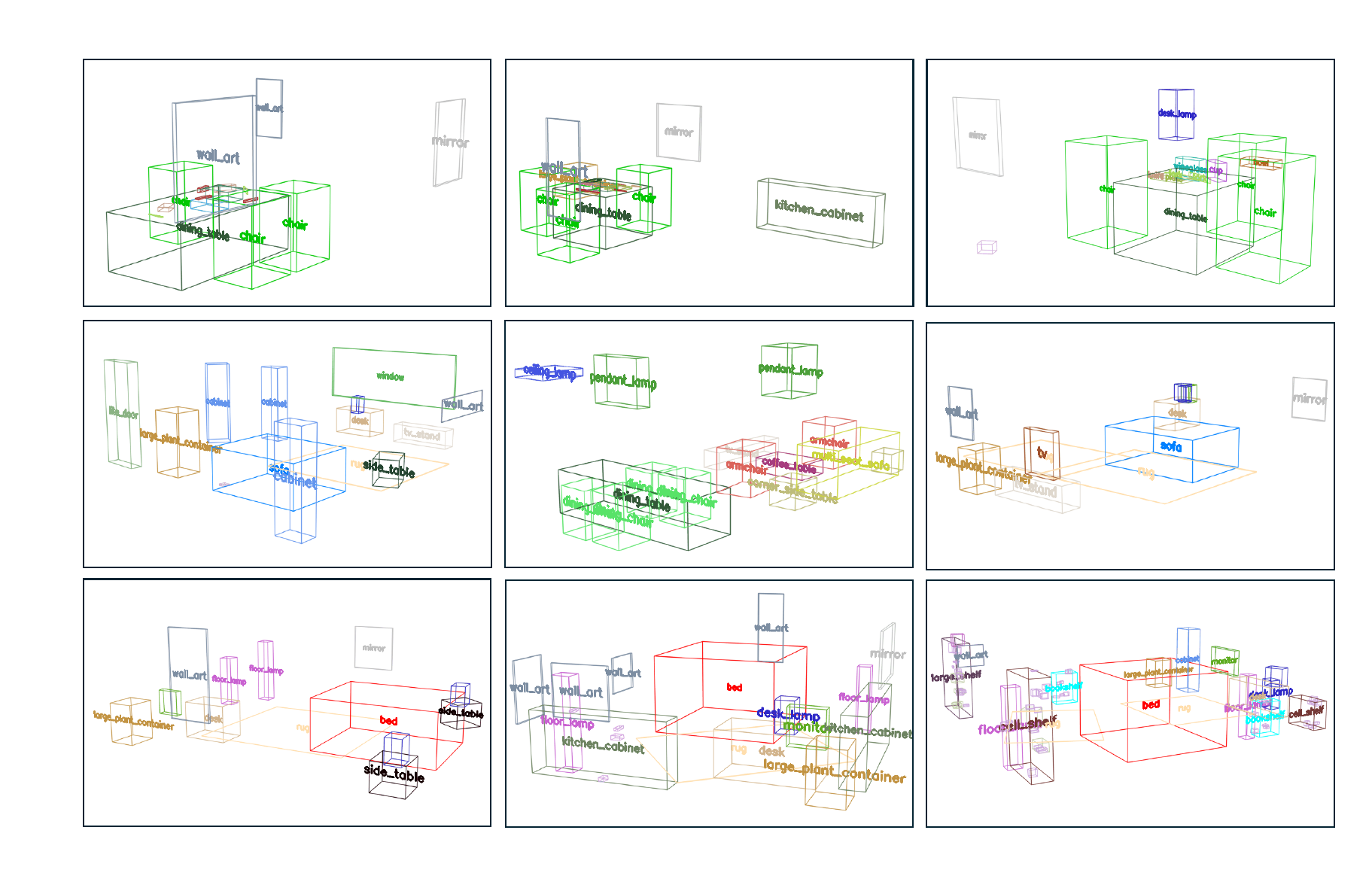}
\caption{Generated layouts by our autoregressive model trained on the M3DLayout dataset, visualized for randomly selected bedroom, living room, and dining room.}
\label{fig:fig_dataset_generation_ar}
\end{figure*}

\begin{figure*}[h]
\centering
\includegraphics[width=1.0\textwidth]{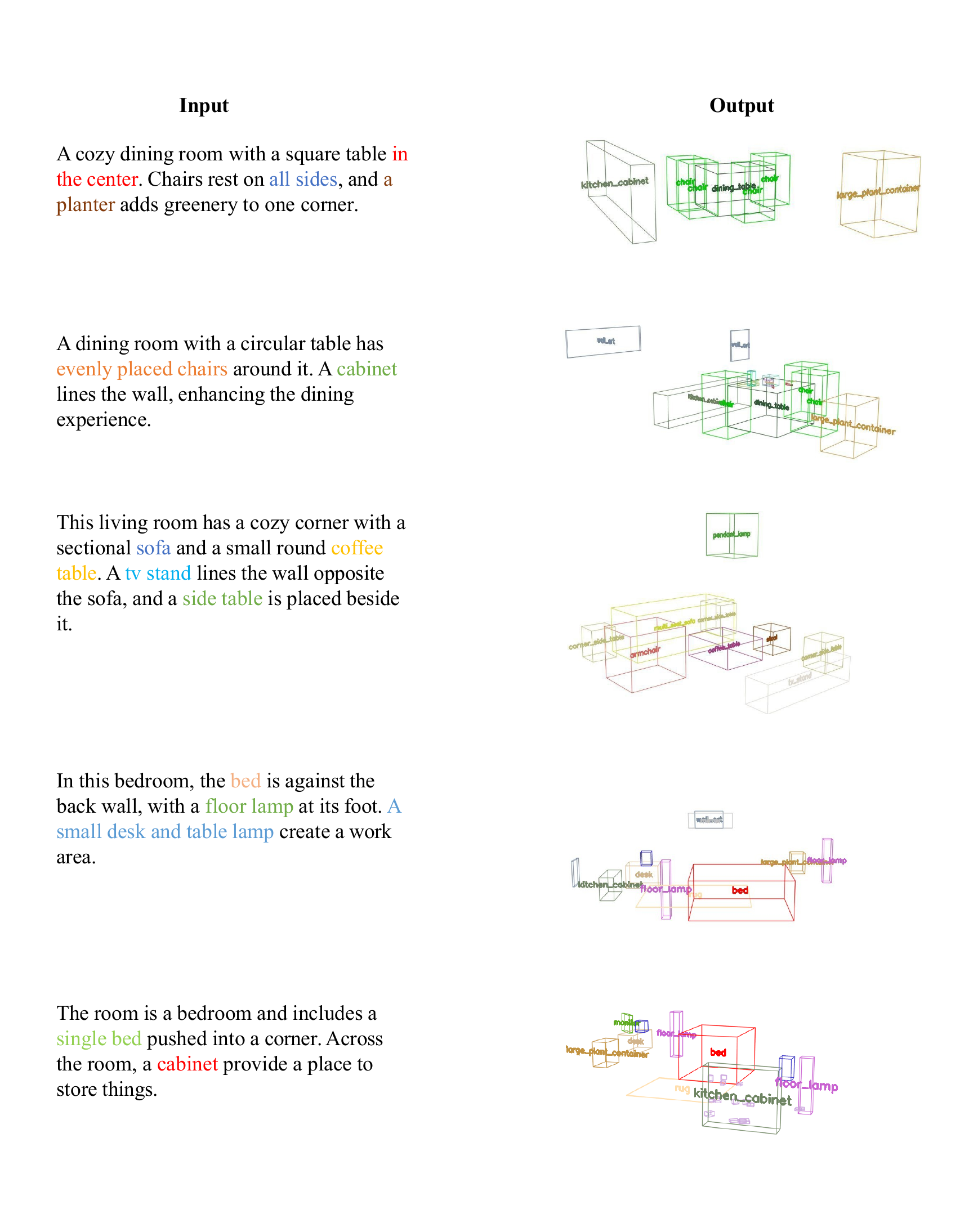}
\caption{Our generated layouts exhibit strong adherence to the provided textual guidance.}
\label{fig:text_context}
\end{figure*}

We visualize more generated layouts by both our diffusion and autoregressive model trained on the M3DLayout dataset, involving bedroom, living room, and dining room in Figure~\ref{fig:fig_dataset_generation} and Figure~\ref{fig:fig_dataset_generation_ar} respectively. From the table, it is evident that our method achieves remarkable performance in both layout coherence and the richness of objects in the generated scenes. We also provide more visualization to indicate strong text coherence of our generated layouts in Figure~\ref{fig:text_context}. These qualitative visualizations further highlight that our method surpasses prior state-of-the-art approaches in both fidelity and controllability.

\section{Ablation Study}
\label{sec:ablation}

\begin{table*}[h]
    \centering
    \caption{\textbf{Ablation studys of diffusion-based methods trained on different datasets.} Lower FID/KID ($\times$0.001) and higher Clip Score indicate better synthesis quality. FID and KID are computed with respect to the real layouts from 3D-FRONT, Matterport, and Inf3DLayout.}
    \resizebox{\textwidth}{!}{%
    \begin{tabular}{ll|ccccccccc}
        \toprule
        \multicolumn{2}{c}{\multirow{2}{*}{Method}} & \multicolumn{3}{c}{\textbf{FID ↓}} & \multicolumn{3}{c}{\textbf{KID ↓}} & \multirow{2}{*}{\textbf{CLIP-Score ↑}} \\
        \cmidrule(lr){3-5}\cmidrule(lr){6-8}
        & & 3D-FRONT & Matterport & Inf3DLayout & 3D-FRONT & Matterport & Inf3DLayout &\\
        \midrule
        \multirow{3}{*}{Diffusion-based} &Ours (3D-FRONT) & 27.33&83.88 &110.98 &10.59 &21.80 &83.45 &0.2083 \\
        &Ours (Matterport) & 81.31& 69.61& 114.58& 46.82& 18.41& 94.45&0.1916\\
        &Ours (Inf3DLayout)  & 93.51&115.07 &54.36 &55.67 &55.53 &34.95 &0.1969 \\
        \midrule
       \multirow{3}{*}{Autoregressive-based} & Ours (3D-FRONT) & 30.02  & 85.08  & 117.39  & 12.09  & 24.46  & 95.11  &  0.2056 \\
       & Ours (Matterport) & 73.36 & 61.08  & 132.54  & 45.11  & 12.58  & 122.21  & 0.1959 \\
       & Ours (Inf3DLayout)  & 105.24  & 122.19  & 60.73  & 70.81  & 65.21 & 43.21 & 0.2026 \\  
        \bottomrule
    \end{tabular}
    }
    \label{tab:ablation}
\end{table*}

We perform the ablation experiments using the diffusion- and autoregressive-based methods to validate the effectiveness of a single training dataset and report the results in Table~\ref{tab:ablation}. As shown in the table, for the diffusion-based method, when the training data and ground truth come from the same dataset, the model trained on a single dataset achieves the best FID and KID compared with the other two models. However, its performance drops significantly when evaluated on data from different datasets. This indicates that, while models trained on a single dataset can effectively fit the distribution of that dataset, they struggle to generalize to varied data. For examplprovided textual guidancee, a model trained on the professional CAD designs dataset (3D-FRONT) encounters difficulties in generating scenes that align with real-world scans (Matterport) or procedurally generation (Inf3DLayout) dataset. Furthermore, the autoregressive-based method exhibits the same characteristics. In contrast, these methods trained on the multi-source M3DLayout dataset achieves balanced performance across data types, producing more realistic and controllable layouts (see Table 3 in Section 5.2 of the present work for comprehensive data).

\section{Object Retrieval}
\label{sec:objectretrieval}
To effectively visualize the generated layouts and meet the evaluation requirements, such as FID (Fréchet Inception Distance), KID (Kernel Inception Distance), and CLIP score, we present a simple, effective and scalable pipeline for layout-to-scene object retrieval.

\begin{figure*}[t]
\centering
\includegraphics[width=1.0\textwidth]{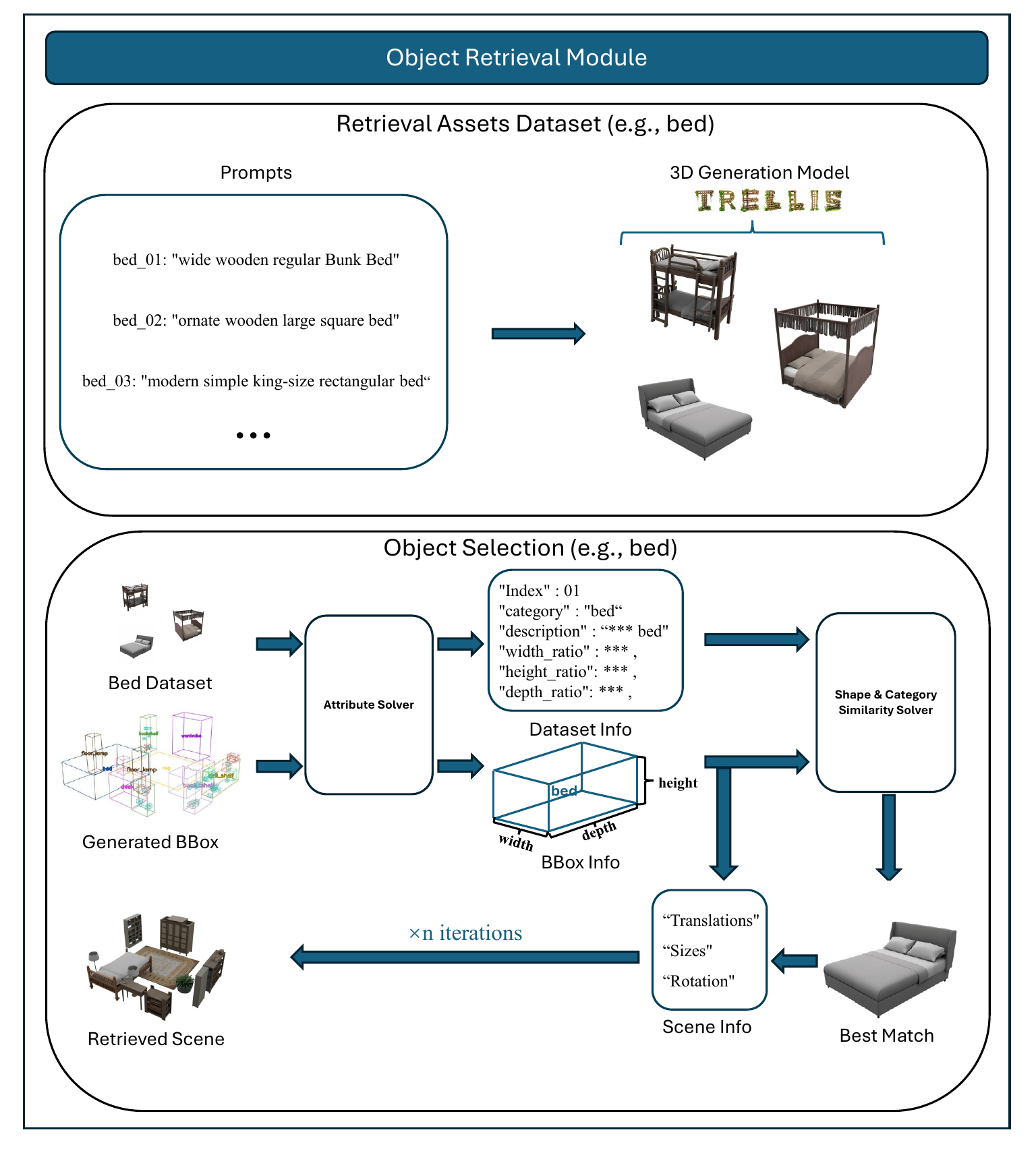}
\caption{\textbf{Retrieval process of layout-to-scene.} This flowchart illustrates the process of retrieving and selecting 3D objects based on generated BBox information. }
\label{fig:object_retrieval_process}
\end{figure*}

In Figure~\ref{fig:object_retrieval_process}, we first build our retrieval dataset by constructing huge amounts of \textit{prompts} for 95 object categories (See details in Table~\ref{tab:retrieval_object_list}), which covers all objects for our dataset, as input for \textit{Text-to-3D generation model}  (TRELLIS~\cite{xiang2025structured}). By delicatly designing our prompts, we can obtain 3D assets with different scales, textures and application scenarios, which can be generalizable to handle with intricate object retrieval process.

In the \textit{Object Selection} phase, the retrieved object, such as a bed, undergoes attribute extraction through the \textit{Attribute Solver}. This solver precomputes attributes like width, height, and depth ratios for each object in the retrieval dataset, and extracts scalar and categorical information from bounding box (BBox) of each object in the generated layout. The BBox, along with additional dataset information, is passed to a \textit{Shape \& Category Similarity Solver} to match the most appropriate object. 

Finally, after iterating all objects in the scene, all best matching objects are chosen and their properties (such as translations, sizes, and rotation) are determined, culminating in the retrieval of the desired 3D scene. This multi-step process ensures accurate retrieval and selection of 3D objects for following applications.

\begin{figure*}[t]
\centering
\includegraphics[width=0.5\textwidth]{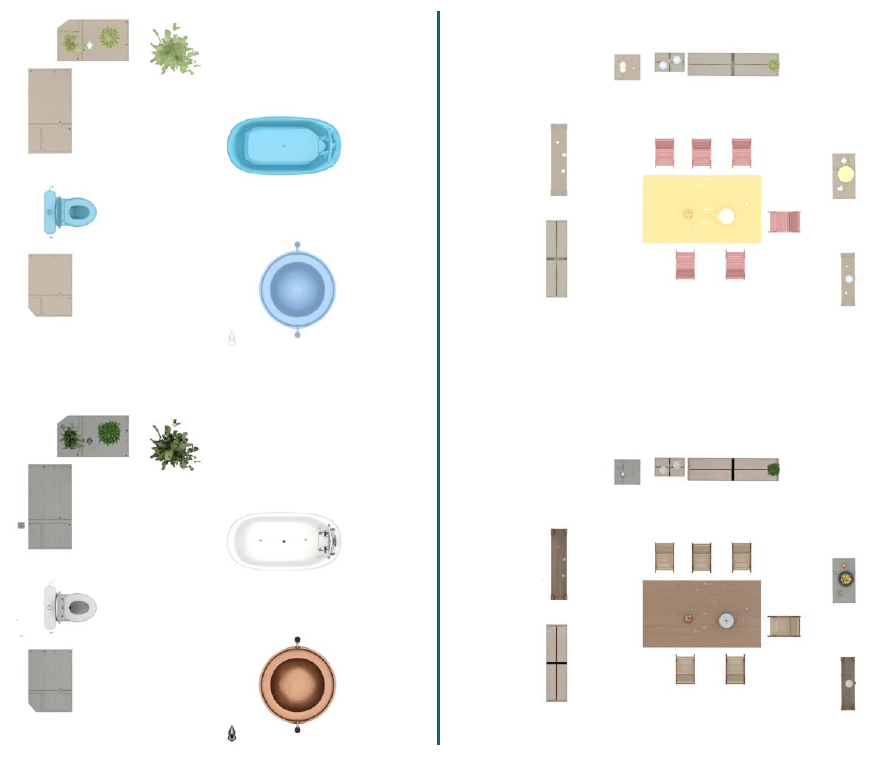}
\caption{\textbf{Retrieval visualization of generated layouts.} The first row displays the retrieved 3D scenes' renderings with pure color schemes. The second row shows the retrieved 3D scenes' renderings with original textures applied.}
\label{fig:object_retrieval_vis}
\end{figure*}

After retrieving successfully, the visualizations provided in this Figure~\ref{fig:object_retrieval_vis} aim to assess both the quality of the generated object retrievals and the performance of the evaluation metrics. The pure color renderings eliminate the influence of textures, making it easier to assess the layout's alignment and object retrieval accuracy using metrics like FID and KID. Meanwhile, the textured renderings offer a visually richer evaluation and applications for users. This approach allows for a comprehensive understanding of the model's effectiveness from both a metric-based and visual quality perspective.

\begin{table*}[ht]
\small
\setlength{\tabcolsep}{6pt}
\renewcommand{\arraystretch}{1.2}

\begin{tabularx}{\textwidth}{|C{0.28\textwidth}|Y|}
\hline
\textbf{Category} & \textbf{Objects (95 in total)}\\
\hline
\textbf{Lighting} &
lighting, ceiling\_lamp, pendant\_lamp, floor\_lamp, desk\_lamp, fan \\
\hline
\textbf{Tables} &
table, coffee\_table, console\_table, corner\_side\_table, round\_end\_table, dining\_table,
dressing\_table, side\_table, nightstand, desk, tv\_stand \\
\hline
\textbf{Seating} &
seating, chair, armchair, lounge\_chair, chinese\_chair, dining\_chair, dressing\_chair,
stool, sofa, loveseat\_sofa, l\_shaped\_sofa, multi\_seat\_sofa \\
\hline
\textbf{Beds} & bed, kids\_bed \\
\hline
\textbf{Shelves \& Book storage} &
shelf, shelving, large\_shelf, cell\_shelf, bookshelf, book, book\_column, book\_stack, nature\_shelf\_trinkets \\
\hline
\textbf{Cabinets \& Wardrobes} &
cabinet, kitchen\_cabinet, children\_cabinet, wardrobe, wine\_cabinet \\
\hline
\textbf{Appliances \& Electronics} &
appliances, microwave, oven, beverage\_fridge, tv, monitor, tv\_monitor \\
\hline
\textbf{Kitchen \& Tableware} &
pan, pot, plate, bowl, cup, bottle, can, jar, wineglass, chopsticks, knife, fork, spoon,
food\_bag, food\_box, fruit\_container \\
\hline
\textbf{Bathroom fixtures} &
bathtub, shower, sink, standing\_sink, toilet, toilet\_paper, toiletry, faucet, towel \\
\hline
\textbf{Doors, Windows \& Coverings} &
glass\_panel\_door, lite\_door, window, blinds, curtain, vent \\
\hline
\textbf{Hardware \& Controls} & hardware, handle, light\_switch \\
\hline
\textbf{Decor} &
plant, large\_plant\_container, plant\_container, vase, wall\_art, picture, mirror, statue,
basket, balloon, cushion, rug, decoration \\
\hline
\textbf{Containers \& Waste} & bag, box, container, clutter, trashcan \\
\hline
\textbf{Architecture \& Elements} & counter, fireplace, pipe, furniture \\
\hline
\textbf{Clothes} & clothes \\
\hline
\textbf{Spaces} & kitchen\_space \\
\hline
\textbf{Gym \& Misc} & gym\_equipment \\
\hline
\end{tabularx}

\caption{\textbf{Category list of retrieval objects.} Our retrieval dataset includes 95 objects which covers nearly all common indoor objects.}
\label{tab:retrieval_object_list}
\end{table*}

{
    \small
    \bibliographystyle{ieeenat_fullname}
    \bibliography{main}
}

\end{document}